\documentclass[sigplan]{acmart}
\usepackage[most]{tcolorbox}
\usepackage{enumitem}
\usepackage{ragged2e}
\usepackage{placeins}
\usepackage{xurl}
\usepackage{hyperref}
\newcommand{\task}[1]{\texttt{\path{#1}}}

\tcbset{
  promptbox/.style={
    width=\linewidth,
    enhanced,
    breakable,
    boxrule=0.4pt,
    arc=2pt,
    left=4pt,
    right=4pt,
    top=4pt,
    bottom=4pt,
    boxsep=2pt,
    before skip=6pt,
    after skip=6pt,
    fonttitle=\bfseries\small,
    fontupper=\small,
    before upper={\RaggedRight\sloppy},
  }
}
\AtBeginDocument{%
  }

\setcopyright{cc}
\setcctype{by}
\acmConference[Agent Skills '26]{Agent Skills '26 Workshop: ACM Conference on AI and Agentic Systems}{May 26, 2026}{San Jos\'e, CA}
\acmBooktitle{Agent Skills '26 Workshop: ACM Conference on AI and Agentic Systems, May 26, 2026, San Jos\'e, CA}
\acmYear{2026}
\acmISBN{}
\acmDOI{}
\begin{document}

\title{
  ClawTrace: Cost-Aware Tracing for LLM Agent Skill Distillation%
}

\author{Boqin Yuan}
\authornote{Both authors contributed equally to this work.}
\affiliation{%
  \institution{University of California San Diego}
  \country{San Diego, CA}
}
\email{b4yuan@ucsd.edu}

\author{Yue Su}
\authornotemark[1]
\authornote{Work done during an internship at Epsilla.}
\affiliation{%
  \institution{Carnegie Mellon University}
  \country{Pittsburgh, PA}
}
\affiliation{%
  \institution{Epsilla}
  \country{Union City, NJ}
}
\email{yuesu@andrew.cmu.edu}

\author{Renchu Song}
\affiliation{%
  \institution{Epsilla}
  \country{Union City, NJ}
}
\email{richard@epsilla.com}

\author{Sen Yang}
\affiliation{%
  \institution{Epsilla}
  \country{Union City, NJ}
}
\email{eric@epsilla.com}

\author{Jing Qin}
\affiliation{%
  \institution{Epsilla}
  \country{Union City, NJ}
}
\email{ricki@epsilla.com}


\begin{abstract}
Skill-distillation pipelines learn reusable rules from LLM agent trajectories, but they lack a key signal: how much each step costs. Without per-step cost, a pipeline cannot distinguish adding a missing step to fix a bug from removing an expensive step that never affected the outcome. We use the cost-attribution gap to ask whether the rule types inside a distilled skill transfer the same way to new tasks. \textbf{ClawTrace} records cost-attributed agent traces and compiles each session into a \textbf{TraceCard}; \textbf{CostCraft} reads TraceCards and writes three kinds of skill patches: \emph{preserve}, \emph{prune}, and \emph{repair}. We find a pattern aggregate metrics hide. On 30 held-out SpreadsheetBench tasks across two seeds, removing prune patches roughly tripled the quality-regression count without lowering median cost. Across the full 84-task SkillsBench transfer, CostCraft saves no aggregate cost. All three quality regressions trace to the preserve lane, and both quality wins trace to the prune lane: prune patches act as quality guardrails while preserve patches drive regressions. We argue that reusable agent skills should be evaluated at the rule-type level, not as monolithic instruction packages. To support this, we release ClawTrace, the TraceCard schema, and the full set of typed skills.
\end{abstract}

%% Single verified CCS concept from the ACM CCS 2012 taxonomy. The
%% concept_id 10010147.10010178 corresponds to "Computing
%% methodologies -> Artificial intelligence" and is stable.






\maketitle
{\let\thefootnote\relax\footnotetext{Code: \url{https://github.com/epsilla-cloud/clawtrace}.\newline Project page: \url{https://www.clawtrace.ai/}.}}
\begin{figure*}[!t]
  \centering
  \includegraphics[width=\textwidth]{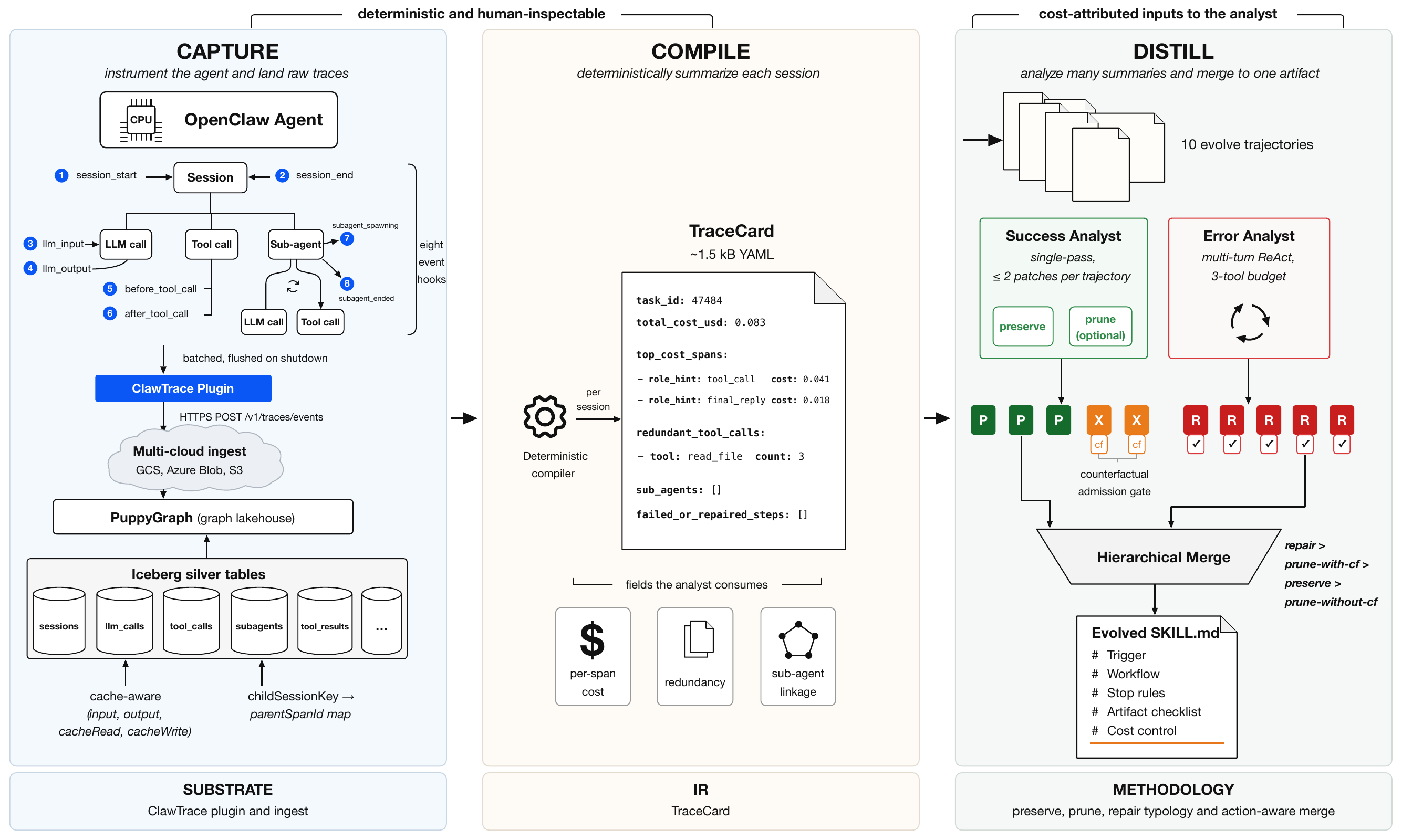}
  \caption{End-to-end architecture. \textbf{Capture}: ClawTrace instruments an agent session via eight event hooks. \textbf{Compile}: a deterministic compiler produces a TraceCard per session. \textbf{Distill}: CostCraft emits preserve, prune, and repair patches that merge into an evolved \texttt{SKILL.md}.}
  \Description{A three-column system diagram showing the ClawTrace and CostCraft pipeline. The left column, Capture, shows an OpenClaw agent emitting eight event hooks for session start and end, LLM input and output, tool calls, and sub-agent lifecycle events. These events are batched by the ClawTrace plugin, posted to a multi-cloud ingest endpoint, and stored in PuppyGraph and Iceberg silver tables. The middle column, Compile, shows a deterministic compiler converting each session into a compact TraceCard containing task id, total cost, top cost spans, redundant tool calls, sub-agent links, and failed or repaired steps. The right column, Distill, shows success and error analysts reading multiple TraceCards, producing preserve, prune, and repair patches, passing prune patches through a counterfactual admission gate, and merging all accepted patches into an evolved SKILL.md file.}
  \label{fig:overview}
\end{figure*}

\section{Introduction}
An agent skill is a structured instruction package that guides an LLM agent without weight updates \cite{schmotz2025agentskillsenablenew}. Recent systems distill such skills directly from agent execution traces \cite{ni2026trace2skilldistilltrajectorylocallessons,zhang2026coevoskillsselfevolvingagentskills,yang2026autoskillexperiencedrivenlifelonglearning}. The aggregate gains are real but uneven: on SkillsBench \cite{li2026skillsbenchbenchmarkingagentskills}, curated skills raise the mean pass rate by 16.2 percentage points, yet 16 of 84 tasks regress and self-generated skills offer no average benefit. Aggregate metrics conceal offsetting wins and losses.

A central reason for these regressions is that existing distillation pipelines treat a skill as a single unit. They partition trajectories into successes and failures, then extract rules from each partition \cite{ni2026trace2skilldistilltrajectorylocallessons}. This partition conflates two distinct operations. \emph{Repairing} a bug requires adding a missing step to a failed trajectory. \emph{Compressing} waste requires removing an expensive but inessential step from a trajectory that already succeeded. The second operation requires per-step cost: the pipeline has to identify which steps were expensive before it can argue that removing them is safe. Standard observability stacks (LangSmith \cite{langsmith2024evaluation}, Langfuse \cite{langfuse2024docs}, Phoenix \cite{phoenix2024llmevals}) record per-span tokens and cost following OpenTelemetry conventions \cite{opentelemetry2024}. They expose this information as dashboard analytics for human operators rather than as a compact summary a distillation pipeline can ingest. A skill-mining analyst needs ranked per-step costs, redundancy flags, and failure markers in a format small enough that many sessions fit a single context window.

This paper contributes on two fronts. First, \textbf{ClawTrace} is an open-source tracing platform that registers eight event hooks covering every LLM call, tool use, and sub-agent spawn. It links child sessions to their parent spans and bills cached tokens at their actual rate. From these events it compiles a \textbf{TraceCard} per session: a $\sim$1.5\,kB YAML summary listing per-step USD cost, typed token counts, redundant tool-call clusters, and failed-or-repaired steps. The plugin currently ships for OpenClaw, and the schema is intentionally framework-agnostic; in principle any agent harness can produce TraceCards by posting conformant JSON to the ingest API, though we have only validated the OpenClaw path end-to-end.

Second, \textbf{CostCraft} reads TraceCards and emits three kinds of skill patches: \emph{preserve} (keep what worked), \emph{prune} (drop unnecessary cost), and \emph{repair} (fix specific failures). The split replaces Trace2Skill's \cite{ni2026trace2skilldistilltrajectorylocallessons} success-versus-error split with a correctness-versus-efficiency split. Section~\ref{sec:costcraft} gives the patch-admission rules and the conflict-aware merge.

The cross-benchmark evaluation supports a third contribution: \emph{distilled agent skills contain separable rule types that transfer in different ways.} On 30 held-out SpreadsheetBench \cite{ma2024spreadsheetbenchchallengingrealworld} tasks across two seeds and the full 84-task SkillsBench \cite{li2026skillsbenchbenchmarkingagentskills} benchmark, prune rules target recurring waste and transferred to new tasks, while preserve rules encoded SpreadsheetBench-specific conventions and overfit. For the agent-skills community, this suggests that reusable skills should be evaluated not only as whole instruction packages, but also by the transfer behavior of their component rule types.

\begin{figure*}[!t]
  \centering
  \includegraphics[width=0.95\linewidth]{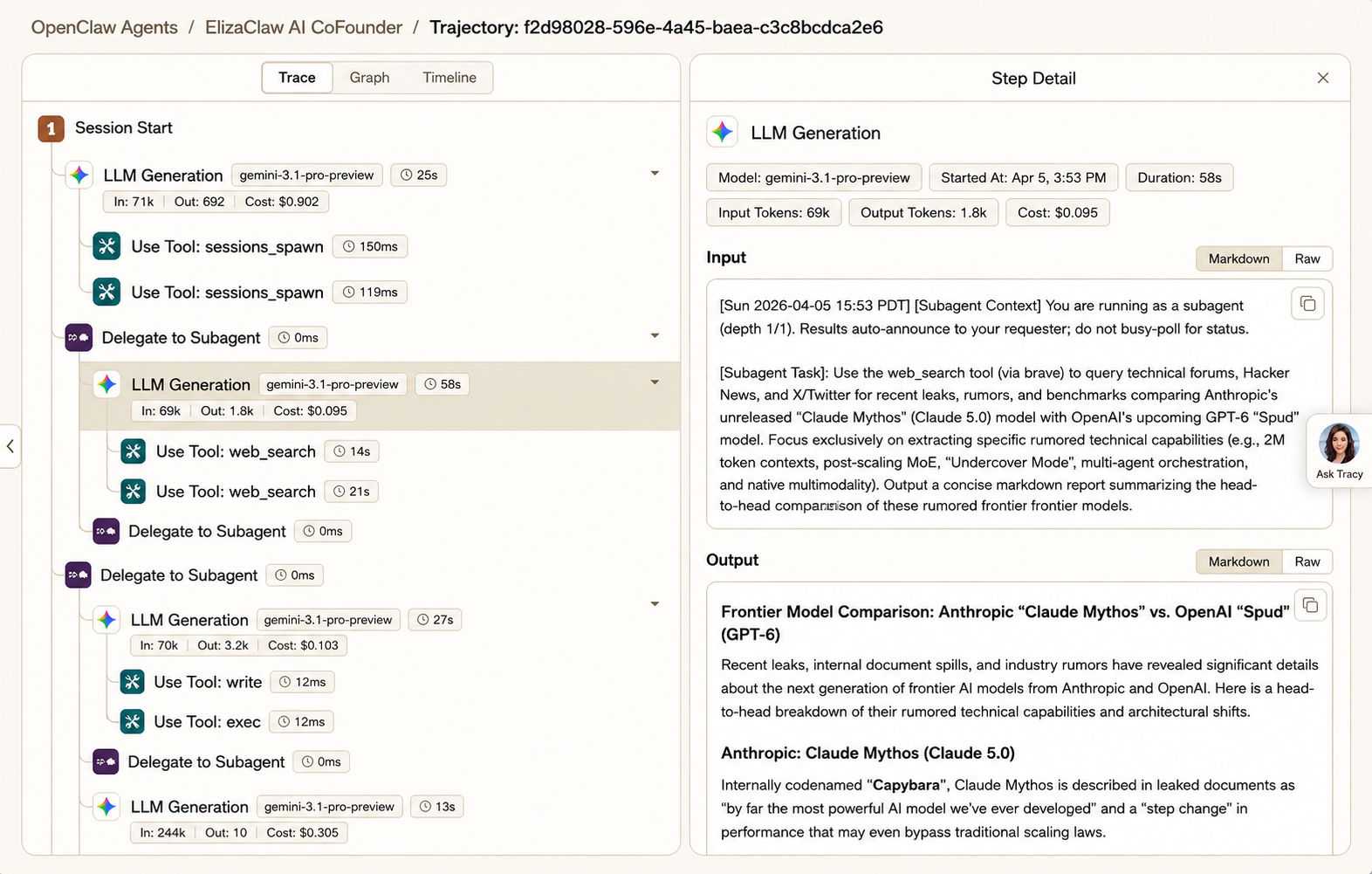}
  \caption{ClawTrace execution-path view showing per-span
  cost attribution, tool-call payloads, and sub-agent nesting
  for a single SpreadsheetBench trajectory.}
  \Description{A screenshot of the ClawTrace interface for one agent trajectory. The left panel lists a nested execution path with session start, LLM generation spans, tool calls, and delegated sub-agent spans. Each span displays model name, duration, token counts, and cost. The selected span is an LLM generation inside a sub-agent. The right panel shows detailed input and output text for that step, including the model, start time, duration, input tokens, output tokens, and dollar cost.}
  \label{fig:clawtrace-screenshot}
\end{figure*}

\section{Related Work}
\label{sec:related}

\paragraph{Agent skills and benchmarks.}
Anthropic \cite{schmotz2025agentskillsenablenew} introduced agent skills as structured packages that augment an LLM agent without weight updates. Xu and Yan \cite{xu2026agentskillslargelanguage} distinguish skills from atomic tools and one-off plans, defining them as reusable modules with applicability conditions. SkillsBench \cite{li2026skillsbenchbenchmarkingagentskills} provides the first systematic benchmark: curated skills raise mean pass rate by 16.2 percentage points, but self-generated skills offer no average benefit. SWE-Skills-Bench \cite{han2026sweskillsbenchagentskillsactually} evaluates skill injection in software-engineering tasks and reports performance drops under context mismatch. SpreadsheetBench \cite{ma2024spreadsheetbenchchallengingrealworld} provides 912 real-world tasks with deterministic cell-match grading and is our primary evaluation benchmark.

\paragraph{Trajectory mining and skill distillation.}
Trace2Skill \cite{ni2026trace2skilldistilltrajectorylocallessons} is the closest prior work. It dispatches parallel analyst sub-agents under a success-versus-error split, with the error analyst using a multi-turn ReAct loop \cite{yao2023reactsynergizingreasoningacting} with oracle access. We adopt its analyst structure but replace the two-way split with a three-way split that adds prune patches for removing waste from successful trajectories. CoEvoSkills \cite{zhang2026coevoskillsselfevolvingagentskills} couples a Skill Generator with a co-evolving Surrogate Verifier, achieving 71.1\% pass rate on SkillsBench through per-task iterative optimization rather than cross-task distillation. EvoSkill \cite{alzubi2026evoskillautomatedskilldiscovery} iteratively diagnoses failures and validates skill updates with ground-truth supervision. AutoSkill \cite{yang2026autoskillexperiencedrivenlifelonglearning} maintains a skill lifecycle from user chat trajectories. SkillWeaver \cite{zheng2025skillweaverwebagentsselfimprove} generates web API skills through structured exploration. These systems address different settings from our offline cross-task distillation.

\paragraph{Experience memory and observability.}
Voyager \cite{wang2023voyageropenendedembodiedagent} accumulates skills through open-ended interaction. Reflexion \cite{shinn2023reflexionlanguageagentsverbal} refines decisions through verbal self-reflection. ReasoningBank \cite{ouyang2026reasoningbankscalingagentselfevolving} stores trajectory-derived insights in a retrieval bank; Trace2Skill shows that distillation into a compact skill document outperforms retrieval by 13.8 percentage points. SkillRL \cite{xia2026skillrlevolvingagentsrecursive} co-evolves skills and policies via reinforcement learning; we study frozen-model, training-free adaptation only. On the observability side, LangSmith, Langfuse, Phoenix, and Arize build on OpenTelemetry \cite{opentelemetry2024} conventions to provide step-level logs and per-span token costs. Dong et al.\ \cite{dong2024agentops} survey the emerging AgentOps landscape and identify cost tracking and multi-agent tracing as open challenges. These tools serve human operators; none produces a compact intermediate representation that a distillation pipeline can consume directly.

\begin{sloppypar}
\paragraph{LLM cost optimization.}
FrugalGPT \cite{chen2023frugalgptuselargelanguage} reduces LLM inference cost through model cascading and caching, and RouteLLM \cite{ong2025routellm} learns to route queries to cheaper models when quality permits. These systems optimize cost at inference time. CostCraft operates at a different point: it mines cost patterns from past trajectories to produce reusable skill rules that reduce waste in future runs.
\end{sloppypar}

\section{Method}
\label{sec:method}

The system has three stages (Figure~\ref{fig:overview}). In the \textbf{Capture} stage, the ClawTrace plugin records raw trace events during an agent session and writes them to cloud storage. In the \textbf{Compile} stage, a deterministic compiler summarizes each session into a TraceCard. In the \textbf{Distill} stage, CostCraft reads TraceCards and produces an evolved skill document through parallel analysis and conflict-aware merging.

\begin{sloppypar}
\subsection{ClawTrace Instrumentation}
\label{sec:clawtrace}

ClawTrace is an OpenClaw-native plugin that registers eight event hooks covering an agent session's full lifecycle: session start and end, LLM input and output, tool-call entry and exit, and sub-agent spawn and termination. Figure~\ref{fig:clawtrace-screenshot} shows the resulting execution-path view with per-span cost attribution. The plugin batches events in memory and flushes them on agent shutdown to a cloud ingest endpoint (\texttt{POST /v1/traces/events}). The endpoint accepts plain JSON with no framework-specific dependency, so any agent harness can produce valid downstream artifacts by posting conformant payloads. A graph-lakehouse pipeline built on PuppyGraph \cite{puppygraph} materializes eight Iceberg silver tables from the raw events. Two design decisions are load-bearing for the rest of the paper.

\emph{Cross-trace linkage for multi-agent runs.} Modern multi-agent systems \cite{wu2023autogen,hong2023metagpt,li2023camel,wang2026lmarslegalmultiagentworkflow} routinely spawn sub-agents that delegate subtasks, producing nested call trees that a flat trace cannot represent. When OpenClaw spawns a sub-agent, ClawTrace links the child session back to the parent's tool-call span through a \texttt{childSessionKey} $\to$ \texttt{parentSpanId} map that persists across flush boundaries. Generic observability stacks support parent-child hierarchies but treat each sub-agent as a separate trace, requiring manual tagging to recover the call graph.

\emph{Cache-aware per-step cost.} ClawTrace bills four token rates separately: \texttt{input}, \texttt{output}, \texttt{cacheRead}, and \texttt{cacheWrite}. Providers price cache reads at roughly one-tenth the fresh input rate. On a typical 50-trajectory SpreadsheetBench run, cache reads constitute 30 to 50 percent of total input volume. Counting them at the fresh-input rate would overstate true cost by 1.6 to 2.0$\times$ and would distort the per-step cost ranking that the distillation pipeline depends on.
\end{sloppypar}
\subsection{TraceCard Compilation}
\label{sec:tracecard}

A TraceCard is a compact YAML summary of one agent session, produced deterministically from the session's span tree. A typical TraceCard is 1.2 to 1.8\,kB, small enough that dozens fit into an analyst's context window. Table~\ref{tab:tracecard} lists the schema fields. Cost and token fields are computed directly from the spans and the provider's pricing table. Three fields use heuristics. \texttt{role\_hint} classifies each LLM step into one of five categories (e.g., \texttt{tool\_call}, \texttt{final\_reply}) based on the step's tool-call ratio and position in the turn sequence. \texttt{redundant\_tool\_calls} groups tool calls with the same name whose arguments are at least 80\% similar by normalized Levenshtein distance, flagging any group of two or more as redundant. \texttt{sub\_agents.output\_used\_in\_final} measures how much of a sub-agent's output appears in the parent's final message using Jaccard overlap.

We audited the redundancy detector on 10 traces (5 with redundant clusters, 5 without). Precision was 100\%: all 8 detected clusters were genuine redundant file reads. Recall was roughly 80\%; one exact-duplicate pair was missed because heredoc payloads drifted past the similarity threshold. The sub-agent Jaccard heuristic is an instrumentation capability that has not been validated on this dataset, because none of the 50 SpreadsheetBench baselines under \texttt{openai-codex/gpt-5.4} spawned sub-agents. Evaluating it requires a multi-agent workload, which we leave to future work.

\begin{table}[t]
\centering
\small
\caption{TraceCard schema. Heuristic fields are marked with $\dagger$.}
\label{tab:tracecard}
\begin{tabular}{@{}lp{0.58\linewidth}@{}}
\toprule
\textbf{Field} & \textbf{Description} \\
\midrule
\texttt{total\_cost\_usd} & Cache-aware USD cost for the session \\
\texttt{total\_tokens} & \texttt{input}, \texttt{output}, \texttt{cacheRead}, \texttt{cacheWrite} \\
\texttt{top\_cost\_spans} & Top-5 spans sorted by \texttt{cost\_usd}, each with \texttt{kind}, \texttt{role\_hint}$^\dagger$, \texttt{tokens}, \texttt{args\_sample} \\
\texttt{redundant\_tool\_calls}$^\dagger$ & Clusters of $\geq$2 calls with Levenshtein $\geq$0.8 \\
\texttt{sub\_agents} & Per-child \texttt{total\_cost\_usd} and \texttt{output\_used\_in\_final}$^\dagger$ (Jaccard) \\
\texttt{failed\_or\_repaired} & Tool results matching error patterns \\
\bottomrule
\end{tabular}
\end{table}

\begin{sloppypar}
\subsection{CostCraft: Three-Action Distillation}
\label{sec:costcraft}

CostCraft is a three-stage distillation pipeline (Figure~\ref{fig:overview}, right panel). Stage~1 runs a set of tasks without any skill and records their trajectories. Stage~2 analyzes each trajectory and proposes skill patches. Stage~3 merges all patches into one skill document. The architecture builds on Trace2Skill's \cite{ni2026trace2skilldistilltrajectorylocallessons} parallel-analyst and hierarchical-merge design; our contributions are the three-way patch typology, the cost-attributed analyst input, and the conflict-aware merge.

\paragraph{Three patch types.} Each patch is labeled with one of three actions. \emph{Preserve} keeps a behavior that contributed to success. \emph{Prune} removes a step that was expensive but did not affect the outcome; it must name the specific high-cost step it targets and include a counterfactual argument for why removal is safe. \emph{Repair} fixes a failure mode found in a failed trajectory, grounded in oracle evidence. The key distinction from prior work is that prune patches come from \emph{successful} trajectories: the task already passed, so the patch is about efficiency, not correctness.

\paragraph{Success Analyst.} This analyst examines each successful trajectory and produces up to two patches: one preserve patch describing the behavior to keep, and optionally one prune patch targeting an expensive step. A prune patch is admitted only when three conditions hold: the analyst names a specific entry from the TraceCard's \texttt{top\_cost\_spans} list, provides a natural-language counterfactual, and phrases the rule as something to avoid rather than a hard cost cap.

\paragraph{Error Analyst.} This analyst examines each failed or partially failed trajectory using a multi-turn ReAct loop with three tools: \texttt{inspect\_mismatches} (read which rubric items failed), \texttt{read\_gold\_snippet} (look up the expected answer), and \texttt{final\_patch} (emit the repair). Oracle access is available only during offline skill authoring, not at agent inference time; the resulting repair rules are distilled into the skill document and applied without oracle access on future tasks. Preserve and prune patches require no oracle at all, since they are mined from naturally successful trajectories. Only repair patches use supervised labels, and only during the offline evolution phase. The analyst has a budget of 3 lookups. If it cannot diagnose the failure within that budget, it emits a low-confidence patch that the merge step deprioritizes.

\paragraph{Conflict-aware merge.} Stage~3 merges all patches into a single skill document via LLM-based hierarchical reduction. The merge enforces a priority order: repair patches with causal diagnosis rank highest, then prune patches with a named cost target and counterfactual, then preserve patches that appear in two or more trajectories. Singleton preserve patches are dropped. When two patches conflict, repair supersedes prune, which supersedes preserve. The final skill has five sections: Trigger, Workflow, Stop rules, Artifact checklist, and Cost control. Post-checks enforce section-heading presence, a 1200-token ceiling, and no task-specific information leakage.
\end{sloppypar}

\subsection{Failure Taxonomy}
\label{sec:failure-taxonomy}

To motivate the three-way patch typology, we built an operational taxonomy from the 16 non-perfect baselines in our 50-task SpreadsheetBench sample, categorizing them into seven failure types (Table~\ref{tab:taxonomy}). This taxonomy is specific to our sample and annotator; we use it to guide analysis, not as a general classification of agent failures. Following Husain \cite{husain2026llmevals}, we coded each baseline by its failure reason, its first three mismatched rubric items, and its TraceCard step-type distribution.

\begin{table}[!htbp]
\centering
\small
\caption{Operational failure taxonomy (16 non-perfect baselines). Each category maps to a patch type.}
\label{tab:taxonomy}
\begin{tabular}{@{}llccl@{}}
\toprule
\textbf{ID} & \textbf{Category} & \textbf{N} & \textbf{\%} & \textbf{Patch} \\
\midrule
T1 & No deliverable & 3 & 19 & repair \\
T2 & Wrong content type & 1 & 6 & repair \\
T3 & Formula not eval. & 1 & 6 & repair \\
T4 & Placeholder mismatch & 4 & 25 & repair \\
T5 & Case/whitespace & 4 & 25 & repair \\
T6 & Logic error & 2 & 13 & preserve \\
T7 & Precision rounding & 1 & 6 & repair \\
\bottomrule
\end{tabular}
\end{table}

Six of seven categories map to repair; one maps to preserve; none maps to prune. This is expected: a failure cannot be fixed by removing a behavior that was never there. Prune patches come only from successful trajectories that contain wasteful steps. The taxonomy also predicts the low prune-match rate in Section~\ref{sec:experiments}: prune rules can only help held-out tasks whose baselines happen to exhibit the same waste patterns observed during training. At our 10-task training scale, the learned prune rules cover two specific wastes (unnecessary workspace-file reads and redundant file re-reads), so held-out tasks without those patterns do not benefit.

\begin{table}[!htbp]
\centering
\small
\caption{Experimental conditions. TC = TraceCard, CF = counterfactual. All conditions share the same merge algorithm and evolve set.}
\label{tab:conditions}
\begin{tabular}{@{}lccc@{}}
\toprule
\textbf{Condition} & \textbf{Cost in TC} & \textbf{Prune} & \textbf{CF gate} \\
\midrule
Baseline ($S_0$) & N/A & N/A & N/A \\
Full CostCraft & \checkmark & merged & enforced \\
No-prune & \checkmark & discarded & N/A \\
No-cost-attr. & stripped & merged & enforced \\
No-CF & \checkmark & merged & disabled \\
\bottomrule
\end{tabular}
\end{table}

\section{Experiments}
\label{sec:experiments}

We ask four questions. (1) Does cost information in TraceCards matter for distillation quality? (2) Do prune patches protect quality, or do they only reduce cost? (3) Do aggregate metrics hide important per-task effects? (4) Does a skill trained on SpreadsheetBench transfer to a different benchmark? All agent runs use \texttt{openai-codex/gpt-5.4} via OpenClaw with ClawTrace instrumentation; we report two seeds (\texttt{seed=0} and \texttt{seed=1}) on the SpreadsheetBench held-out set unless noted otherwise.

\begin{sloppypar}
\subsection{Setup}
\label{sec:setup}

We evaluate on SpreadsheetBench \cite{ma2024spreadsheetbenchchallengingrealworld}, a 912-task benchmark where the agent must produce a spreadsheet and a deterministic grader checks each cell against the gold answer. Per-task quality $Q \in [0,1]$ is the fraction of rubric items satisfied. $Q{=}1.0$ means every cell matches; $Q{=}0$ means no scorable file was produced. Per-task cost is the cache-aware USD total from ClawTrace. We sanitize the agent workspace before each run to prevent earlier runs from leaking learned context into later ones.

The experiment has four phases. First, we run 50 tasks sampled from the 200-task professional subset, stratified by difficulty, with no skill and record each trajectory. Second, we split the 50 tasks into an evolve set of 10 for training, a held-out set of 30 for evaluation, and 10 reserved for pipeline development. The evolve set contains 4 successes, 4 partial successes, and 2 failures. Third, we run CostCraft on the 10 evolve-set TraceCards to produce one skill. We repeat this step under four ablation conditions (Table~\ref{tab:conditions}), each time removing one pipeline signal to produce a different skill. Fourth, we run all 30 held-out tasks under each skill at the same seed as the original baseline and compare quality and cost per task.

The ablation conditions map onto prior work. No-cost-attribution approximates the Trace2Skill \cite{ni2026trace2skilldistilltrajectorylocallessons} information regime: the analyst sees trajectory structure and outcomes but not per-step cost. No-prune approximates a success/failure-only patch policy where successful trajectories yield only preserve rules.

Skill distillation always risks regressions: rules learned from one task can conflict with the requirements of another. SkillsBench \cite{li2026skillsbenchbenchmarkingagentskills} reports this pattern even for human-curated skills (16 of 84 tasks get worse). Our evaluation measures whether each pipeline signal reduces the regression rate, not whether distillation eliminates regressions entirely.
\end{sloppypar}

\begin{sloppypar}

\subsection{Main Results}
\label{sec:main-results}

\begin{figure*}[t]
  \centering
  \begin{minipage}[t]{0.48\textwidth}
    \centering
    \includegraphics[width=\linewidth]{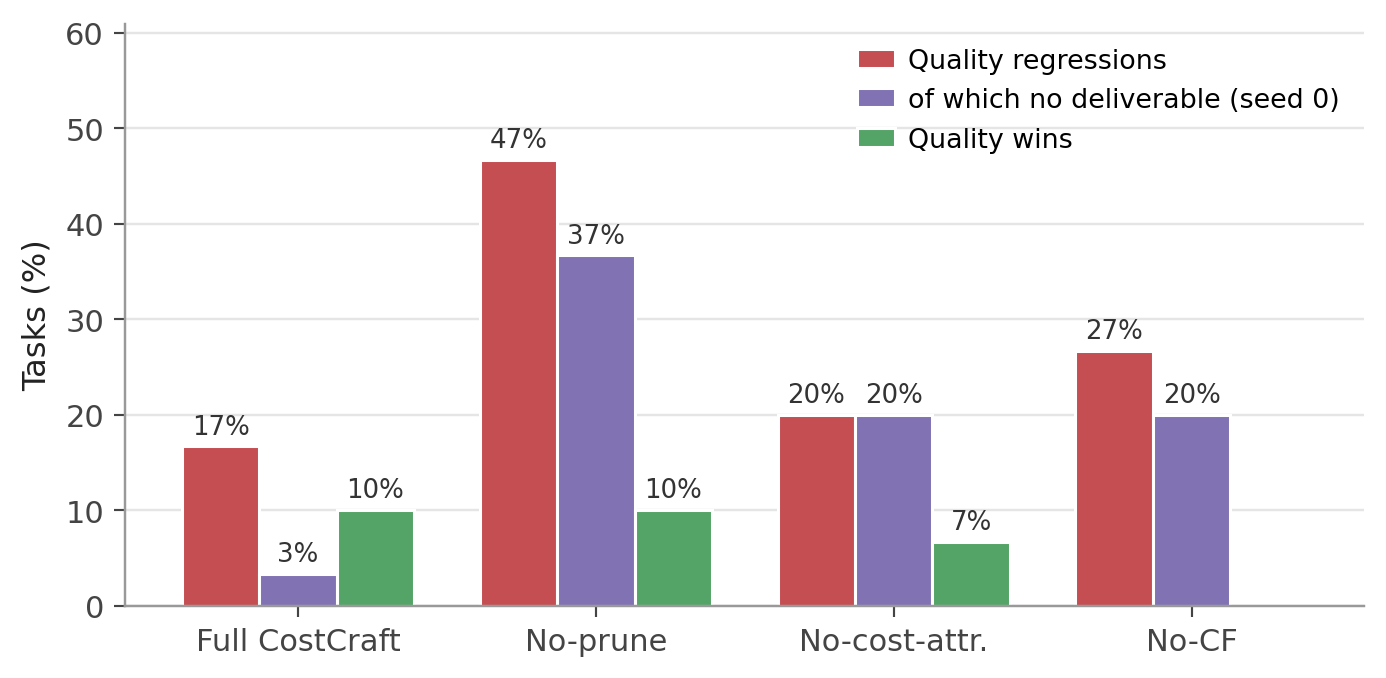}
    \caption{Quality outcome rates across ablation conditions on 30 held-out SpreadsheetBench tasks. Regressions and wins for Full CostCraft and No-prune are pooled per-task means across two seeds; No-cost-attribution and No-CF report a single seed ($N{=}30$ and $N{=}15$). The middle (purple) bar in each group is the seed-0 subset of regressions that delivered $Q{=}0$ from a passing baseline; this severity is seed-sensitive (Appendix~\ref{app:multiseed}). Full CostCraft has the lowest regression rate ($17\%$) and matches No-prune on win rate ($10\%$).}
    \Description{A grouped bar chart comparing four CostCraft conditions: Full CostCraft, No-prune, No-cost-attribution, and No-counterfactual. For each condition, the chart reports the percentage of tasks with quality regressions, the seed-0 subset of regressions that delivered no output from a passing baseline, and quality wins. Full CostCraft has the lowest regression rate at 17 percent. No-prune has the largest regression rate at 47 percent and the largest no-deliverable rate at 37 percent. No-cost-attribution and No-CF show 20 and 27 percent regression rates respectively.}
    \label{fig:ablation}
  \end{minipage}%
  \hfill
  \begin{minipage}[t]{0.48\textwidth}
    \centering
    \includegraphics[width=\linewidth]{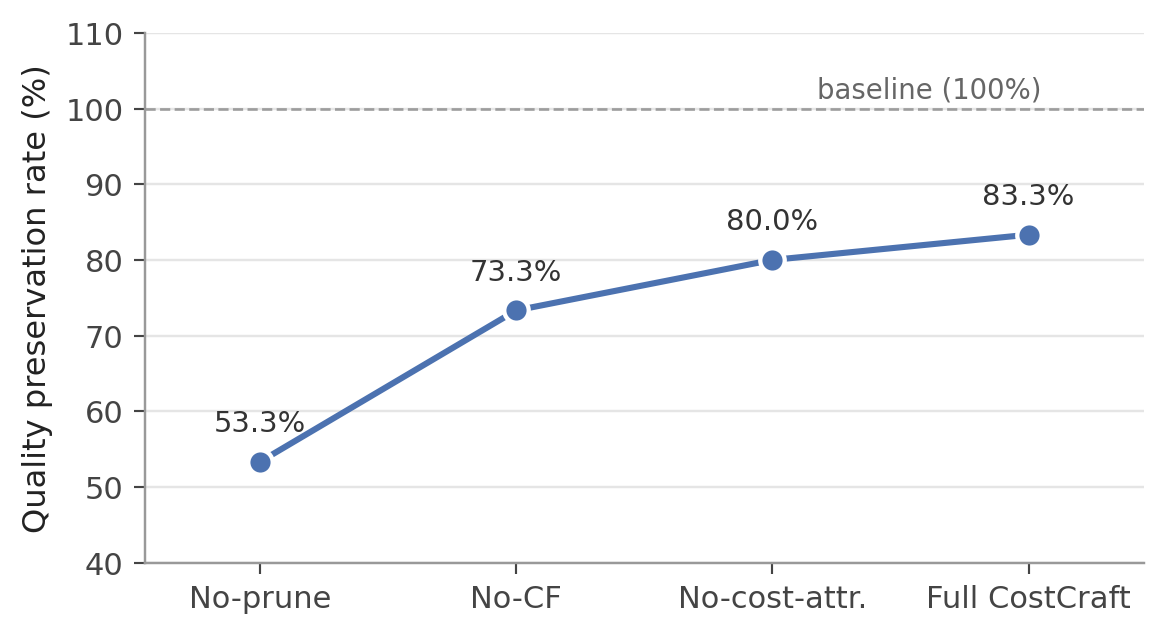}
    \caption{Quality preservation rate across ablation conditions. Full CostCraft and No-prune use pooled per-task means across two seeds; No-cost-attribution and No-CF use a single seed. Each point removes one pipeline signal (left to right adds signal). Full CostCraft preserves quality on $83.3\%$; No-prune drops to $53.3\%$.}
    \Description{A line chart showing quality preservation rate across ablation conditions. The x-axis orders conditions from more signal removed to the full pipeline: No-prune, No-counterfactual, No-cost-attribution, and Full CostCraft. The quality preservation rate increases from 53.3 percent for No-prune, to 73.3 percent for No-counterfactual, to 80.0 percent for No-cost-attribution, and to 83.3 percent for Full CostCraft. A dashed horizontal line marks the 100 percent baseline reference.}
    \label{fig:signal}
  \end{minipage}
\end{figure*}

Each held-out task is compared to its own baseline under the same seed. A task regresses when $Q_{\mathrm{skill}} < Q_{\mathrm{baseline}} - 0.01$; a win is the reverse. Figure~\ref{fig:ablation} reports the outcomes.

\emph{Cost attribution and prune rules each carry independent weight.} Stripping cost from TraceCards (the No-cost-attribution condition) more than doubles the median cost uplift on successful tasks, from $+22\%$ to $+49\%$, and raises the regression count from 4 to 6. Five of those six are catastrophic, with the agent stopping at $Q{=}0$ without producing output; Full CostCraft exhibits this failure mode only once. Independently, removing prune patches (the No-prune condition) triples the regression count at seed 0 from 4 to 13 while median cost stays comparable ($+15\%$ vs.\ $+21\%$), and eight of the 13 No-prune regressions produce no output at all. The two ablations alter different pipeline signals, so the effects can be attributed separately to cost attribution and to prune-rule presence.

The prune-derived Cost-control section thus functions as a quality guardrail in this ablation rather than as a pure cost-compression mechanism. Without it, the remaining skill sections push the agent toward broken behavior on tasks that already worked: the agent continues issuing tool calls but never writes the final answer. This protective role is orthogonal to cost compression and is not captured by Trace2Skill's \cite{ni2026trace2skilldistilltrajectorylocallessons} success-versus-error split.

\emph{The regression-count gap is robust across seeds; severity is not.} Repeating all 30 held-out tasks at \texttt{seed=1} preserves the direction: Full CC has 3 regressions versus No-prune's 5, and pooled per-task means across both seeds show 5 versus 14 regressions ($2.8\times$). The ``no deliverable'' severity is more seed-sensitive: 11 of 13 No-prune regressions deliver $Q{=}0$ at seed 0, but only 1 of 5 at seed 1. Within-seed cost has a coefficient of variation near 43\% on this benchmark, so we report cost percentages as approximate; per-seed and pooled tables are in Appendix~\ref{app:multiseed}.

\emph{Aggregation hides per-task effects.} All three skill conditions show a positive median cost change, yet the same conditions contain two full recoveries ($Q{=}0 \to 1.0$ and $Q{=}0 \to 0.84$) alongside catastrophic No-prune failures. Without a per-task breakdown, these wins and losses cancel and the result reads as noise.

\end{sloppypar}

\subsection{Regime-Partitioned Breakdown}
\label{sec:regime}

Figure~\ref{fig:signal} shows the cumulative effect: each ablation removes one pipeline signal, and quality preservation drops monotonically from $83.3\%$ under Full CostCraft to $53.3\%$ under No-prune. Splitting the 30 held-out tasks by their baseline outcome reveals effects that aggregate numbers cancel out (Table~\ref{tab:regime} in Appendix~\ref{app:regime}). On failed baselines, repair patches recover 2 of 3 tasks to near-perfect quality, at a cost premium that failed tasks can tolerate. On successful baselines, 2 of 17 tasks contain the specific waste patterns targeted by the learned prune rules. On those 2 tasks, Full CostCraft still increases cost (median +30\%), but the No-prune ablation causes one of them to fail entirely ($Q{=}0$).

\subsection{Prune Coverage and the Efficiency Lever}
\label{sec:prune-coverage}

The trained skill contains two prune rules, each learned from 3 training trajectories: (1)~skip workspace-memory files (\texttt{MEMORY.md}, \texttt{SOUL.md}) when the task is self-contained, and (2)~read each input file once and cache its content rather than re-reading.

A held-out task matches a prune rule when its baseline TraceCard contains the targeted waste pattern. In our 30-task sample, 2 of 17 successful held-out tasks match (11.8\%). On those 2 tasks, Full CostCraft still increases cost (+38\% and +22\%), but the No-prune ablation causes one to fail entirely ($Q{=}0$). In this sample, the prune rules are preventing quality collapse rather than compressing cost; whether this protective role generalizes beyond our training scale remains open.

\paragraph{Counterfactual gate (No-CF).} On a 15-task success-heavy subset, disabling the counterfactual admission gate produced 2 additional regressions compared to Full CostCraft, even though the analyst voluntarily included counterfactual text in all prune patches regardless of the gate setting. The gate appears to shape counterfactual \emph{quality} rather than \emph{presence}. The sample is small ($N{=}15$); we treat this as suggestive (Appendix~\ref{app:cf}).

Three per-task case studies (recovery, protection, and regression) are in Appendix~\ref{app:cases}.

\subsection{SkillsBench Cross-Benchmark Evaluation}
\label{sec:skillsbench}

\begin{figure*}[t]
  \centering
  \includegraphics[width=\textwidth]{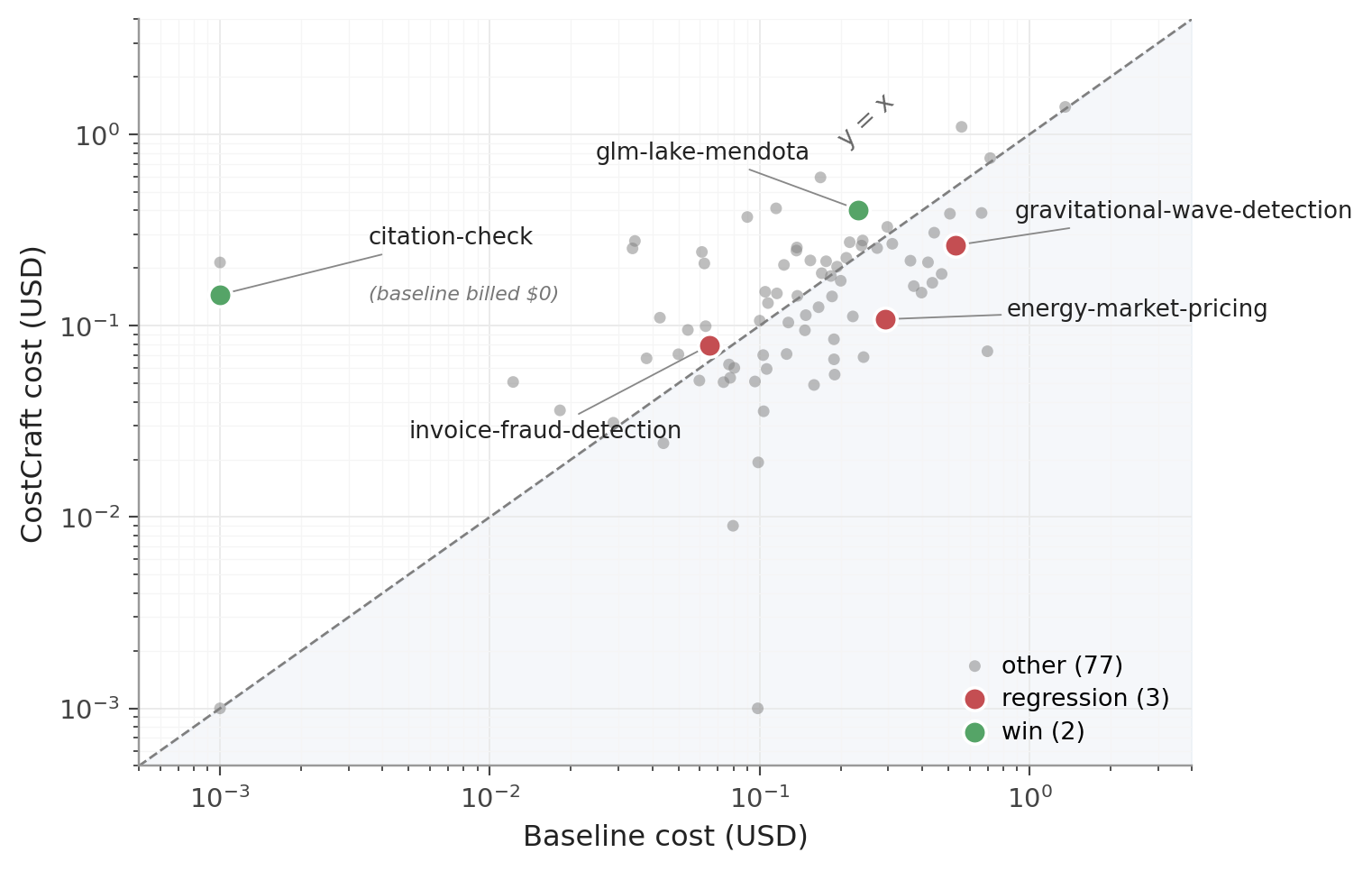}
  \caption{Per-task cost on the full 84-task SkillsBench benchmark (log--log axes; 82 cost-paired tasks shown after Docker-eval failures). Each marker is one task. Below the dashed $y{=}x$ reference, CostCraft is cheaper than the baseline; above, more expensive. Quality outcomes are colored: green for the two wins, red for the three regressions; all other tasks are gray. The wins and regressions are the same five events enumerated in Table~\ref{tab:skillsbench-events}. \texttt{citation-check} sits at the left edge because the baseline billed no cost (clamped to \$$10^{-3}$ for log display).}
  \Description{A log-log scatter plot of per-task cost on the full 84-task SkillsBench benchmark. The x-axis is baseline cost in USD and the y-axis is CostCraft cost in USD, both on a logarithmic scale from one tenth of a cent to four dollars. A dashed diagonal line marks y equals x. Most tasks form a gray cloud near the diagonal. Three red markers labeled energy-market-pricing, gravitational-wave-detection, and invoice-fraud-detection show the three regressions. Two green markers labeled citation-check and glm-lake-mendota show the two wins; citation-check is at the left edge of the plot because its baseline cost was zero. A subtle blue shaded region marks the area below the diagonal where CostCraft is cheaper than the baseline.}
  \label{fig:skillsbench}
\end{figure*}

To test whether a SpreadsheetBench-trained skill transfers to other task types, we ran the full SkillsBench \cite{li2026skillsbenchbenchmarkingagentskills} benchmark (84 tasks) under two conditions: baseline (\texttt{gpt-5.4} with no skill) and baseline plus the SpreadsheetBench-trained CostCraft skill. The tasks cover data processing, scientific computing, document analysis, security, code generation, video processing, and agent planning; none involves spreadsheets. Each task is graded by a Docker-contained pytest verifier that emits a binary pass/fail. Three N values appear in what follows. Of the 84 tasks, 4 failed at the Docker-eval container layer on both conditions (image build errors or test timeouts) and contribute no usable signal. The remaining tasks yield \textbf{82 cost-paired tasks} from which Figure~\ref{fig:skillsbench} is drawn, \textbf{74 quality-gradeable pairs} on which we report wins, regressions, and ties (8 had build or grading issues on at least one side), and \textbf{79 paired-direction tasks} for the cost-decrease direction statistic in the next paragraph (computed only on tasks where the baseline billed nonzero cost). ClawTrace's TraceCard compiler ran on every trace without modification.

CostCraft does not save cost on aggregate over this benchmark: the median baseline cost is \$0.143 and the median CostCraft cost is \$0.144, indistinguishable at the precision the experiment supports. Many SkillsBench tasks fail under both systems, so aggregate cost mixes successful and no-deliverable trajectories and is dominated by domain-hard items such as manufacturing scheduling, PDDL planning, and quantum simulation, on which the baseline returns no deliverable and CostCraft neither helps nor hurts. We therefore treat paired direction and trace-attributed failure mode as the primary transfer signal rather than the aggregate. Cost decreased on 41 of 79 paired tasks (52\%), with a median $-45\%$ reduction where it decreased and a median $+43\%$ increase where it increased. Figure~\ref{fig:skillsbench} visualizes the full-benchmark cost pattern, with the five quality-changing events highlighted.

\begin{sloppypar}
\paragraph{Prune rules transferred; preserve rules overfit.} Each of the five quality-changing events on the full benchmark is attributable, by inspection of the merged skill, to a single patch lane (Table~\ref{tab:skillsbench-events}). The three regressions trace to the preserve lane: its SpreadsheetBench-specific formatting rules inject conventions that the SkillsBench verifier does not expect. Two of the three also exhibit large cost drops ($-63\%$, $-50\%$) because the agent follows the bad rule but does so cheaply, never reaching the work the verifier scores. The two wins trace to the prune lane: the cost-control rules free enough session headroom for the agent to reach a final output step that the baseline either fails to attempt (\texttt{citation-check} bills no cost on the baseline) or attempts at lower fidelity. Quality recovery comes at a cost premium on \texttt{glm-lake-mendota} ($+74\%$); the per-call cost reduction from the ``read each file once and cache'' rule shows up not on the wins but on the broader 52\% of paired tasks where cost decreased.

\begin{table}[t]
\centering
\footnotesize
\setlength{\tabcolsep}{4pt}
\caption{Trace-attributed SkillsBench events. The three regressions trace to the preserve lane (SpreadsheetBench-specific formatting rules); the two wins trace to the prune lane (cost-control rules freeing session headroom). ``---'' marks an undefined ratio (baseline bills no cost).}
\label{tab:skillsbench-events}
\begin{tabular}{@{}lccr@{}}
\toprule
\textbf{Task} & \textbf{Outcome} & \textbf{$Q$} & \textbf{$\Delta$\,cost} \\
\midrule
\multicolumn{4}{@{}l}{\emph{Regressions (preserve lane)}} \\
\texttt{energy-market-pricing}        & reg. & $1.0\!\to\!0.0$ & $-63\%$ \\
\texttt{gravitational-wave-detection} & reg. & $1.0\!\to\!0.0$ & $-50\%$ \\
\texttt{invoice-fraud-detection}      & reg. & $1.0\!\to\!0.0$ & $+21\%$ \\
\midrule
\multicolumn{4}{@{}l}{\emph{Wins (prune lane)}} \\
\texttt{citation-check}               & win  & $0.0\!\to\!1.0$ & ---     \\
\texttt{glm-lake-mendota}             & win  & $0.0\!\to\!1.0$ & $+74\%$ \\
\bottomrule
\end{tabular}
\end{table}

The asymmetry is the central cross-benchmark observation. At full-benchmark scale, the regression-to-win ratio is 3:2, and the lane attribution is clean rather than mixed: every regression is a preserve-lane case and every win is a prune-lane case. We therefore treat the 3:2 figure not as five stochastic outcomes but as five trace-attributed mechanism instances, each documented by an explicit rule conflict or rule firing. Separating the two patch types at distillation time appears important for cross-task generalization; we treat the ratio itself as a directional signal rather than a stable rate estimate, since larger samples would be needed to place a confidence interval on it.

Tracing overhead is bounded at ${\approx}$0.30\% of agent wall time with zero quality divergence on 10 paired ON/OFF reruns (Appendix~\ref{app:overhead}).
\end{sloppypar}
\section{Limitations}
\label{sec:limitations}

The claims in this paper are scoped to a single agent runtime: \textbf{OpenClaw with the \texttt{openai-codex/gpt-5.4} backbone}. Every result reported in the abstract, §4, and §6 is for that combination; we make no claim about Claude Code \cite{anthropic2025claudecode}, AutoGen \cite{wu2023autogen}, MetaGPT \cite{hong2023metagpt}, or any other agent runtime, and we make no claim about other LLM backbones such as Claude, Gemini, or open-weights models. Whether the rule-type transfer asymmetry generalizes beyond this single agent--backbone pair is open. Within this scope, we evaluate 30 SpreadsheetBench held-out tasks at two seeds (60 runs per condition) and 74 quality-gradeable SkillsBench pairs out of 84. Within-seed cost has a coefficient of variation near 43\% on this benchmark, so cost percentages are calibrated to two-seed pooled means and a third seed would not tighten them meaningfully. The catastrophic No-prune severity in Section~\ref{sec:main-results} (deliveries of $Q=0$ on tasks where the baseline succeeded) is seed-0-specific: the regression \emph{count} gap between Full CC and No-prune is preserved at seed 1, but the severity collapses (one $Q=0$ run at seed 1 versus eleven at seed 0). The failure taxonomy was coded by one annotator; a second annotator would be needed to measure inter-coder reliability. Two of the three heuristic TraceCard fields (\texttt{sub\_agents.output\_used\_in\_final} and \texttt{failed\_or\_repaired\_steps}) were not exercised because the tested backbone did not spawn sub-agents.

At the 10-task training scale, two prune rules emerge and match 2 of 17 successful held-out tasks; a richer prune library would require a larger training set. The SkillsBench evaluation tests cross-benchmark transfer but does not repeat the full ablation matrix (No-prune, No-cost-attribution) on SkillsBench; that ablation is the natural next step. ClawTrace currently supports OpenClaw; portability to other agent harnesses through the ingest API has not been validated end-to-end. We treat these boundaries as the scope of the present claims, not as fragility. The empirical levers most likely to extend the rule-type transfer story beyond this perimeter are multi-backbone evaluation, multi-runtime evaluation (Claude Code, AutoGen, others), a larger prune library, and a SkillsBench ablation matrix.

\begin{sloppypar}

\section{Conclusion}
\label{sec:conclusion}

Distilled agent skills should be evaluated at the rule-type level, not as monolithic instruction packages, because their component rule types transfer in qualitatively different ways. Two pieces of evidence carry this argument. Within-distribution, removing prune patches roughly tripled the regression count on SpreadsheetBench across two seeds, while median cost stayed unchanged: prune rules at training scale act as a quality guardrail. Cross-distribution, on the full 84-task SkillsBench transfer, all three quality regressions trace to the preserve lane and both quality wins trace to the prune lane (Table~\ref{tab:skillsbench-events}); the same lanes drive cost decreases on roughly half of paired tasks. The asymmetry is what aggregate metrics hide.

We release ClawTrace, all TraceCards, and evolved skills at \url{https://github.com/epsilla-cloud/clawtrace}; the project page is at \url{https://www.clawtrace.ai/}. All evidence reported here is for the OpenClaw runtime with the \texttt{gpt-5.4} backbone. Whether the pattern generalizes to other runtimes or other backbones is open, and is the natural next direction together with a larger evolve set and a SkillsBench ablation matrix.
\end{sloppypar}

\bibliographystyle{ACM-Reference-Format}
\bibliography{sample-base}

%% Flush any pending floats so appendix figures cannot drift up into
%% the bibliography page.
\FloatBarrier

\appendix
\section*{Appendix}
\addcontentsline{toc}{section}{Appendix}
\renewcommand{\thesubsection}{\Alph{subsection}}

\subsection{ClawTrace Platform Demo}
\label{app:platform}

ClawTrace is an observability and optimization platform for OpenClaw agents. Its goal is to make agents better, cheaper, and faster by giving operators real-time tracing, cost analysis, and actionable recommendations. The platform provides four synchronized views over the same span tree, each serving a different stage of the diagnosis workflow.

The execution-path view (Figure~\ref{fig:clawtrace-screenshot} in the main text) renders the full call graph of an agent session as an interactive trace tree. Every LLM call, tool use, and sub-agent delegation is visible with its payload, token breakdown, and USD cost. This is the view the CostCraft analyst consumes through the TraceCard abstraction.

\begin{figure}[!htbp]
  \centering
  \includegraphics[width=0.95\linewidth]{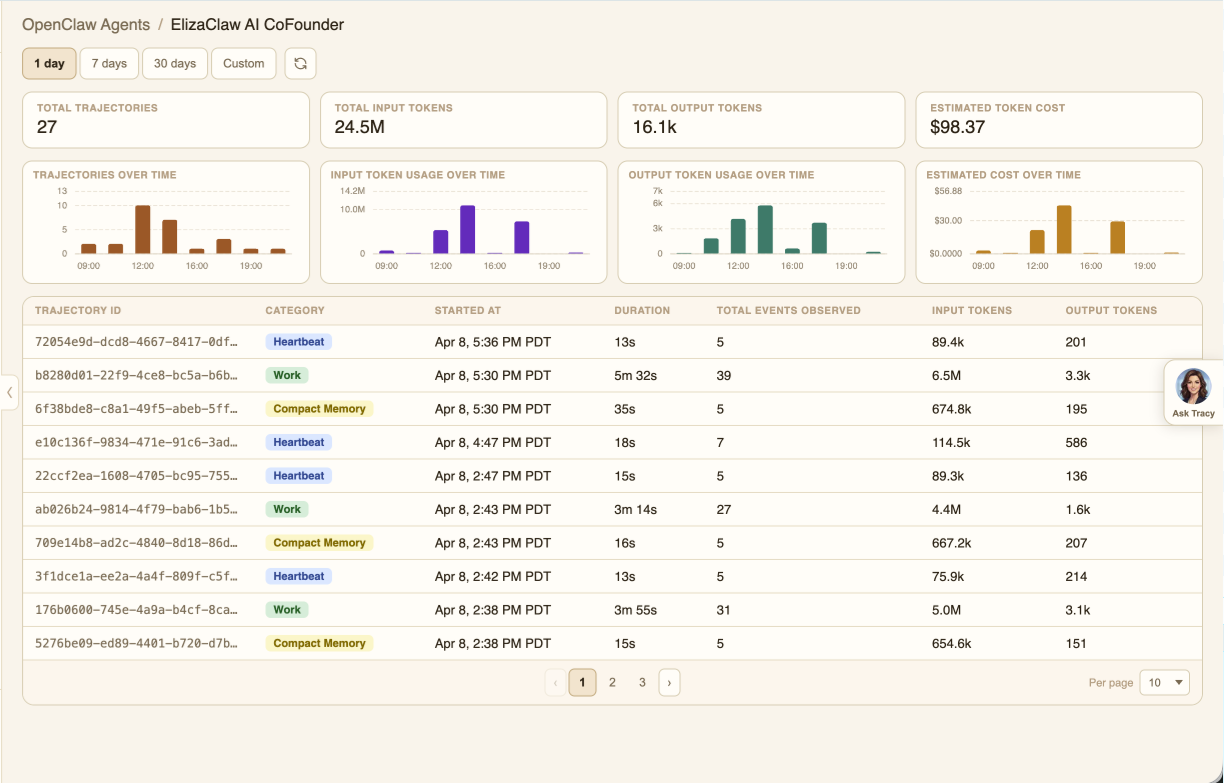}
  \caption{Trajectory dashboard. Each row is one agent run, with
  columns for total cost, token count, step count, and outcome.
  Daily trend charts and per-agent filtering let operators spot
  cost spikes before drilling into individual traces.}
  \Description{A ClawTrace trajectory dashboard for one OpenClaw agent. The top row summarizes total trajectories, total input tokens, total output tokens, and estimated token cost. Four small charts show trajectories over time, input token usage over time, output token usage over time, and estimated cost over time. A table below lists individual trajectories with truncated trajectory IDs, category labels such as Heartbeat, Work, and Compact Memory, start time, duration, total events observed, input tokens, and output tokens.}
  \label{fig:dashboard}
\end{figure}

\begin{figure}[!htbp]
  \centering
  \includegraphics[width=0.95\linewidth]{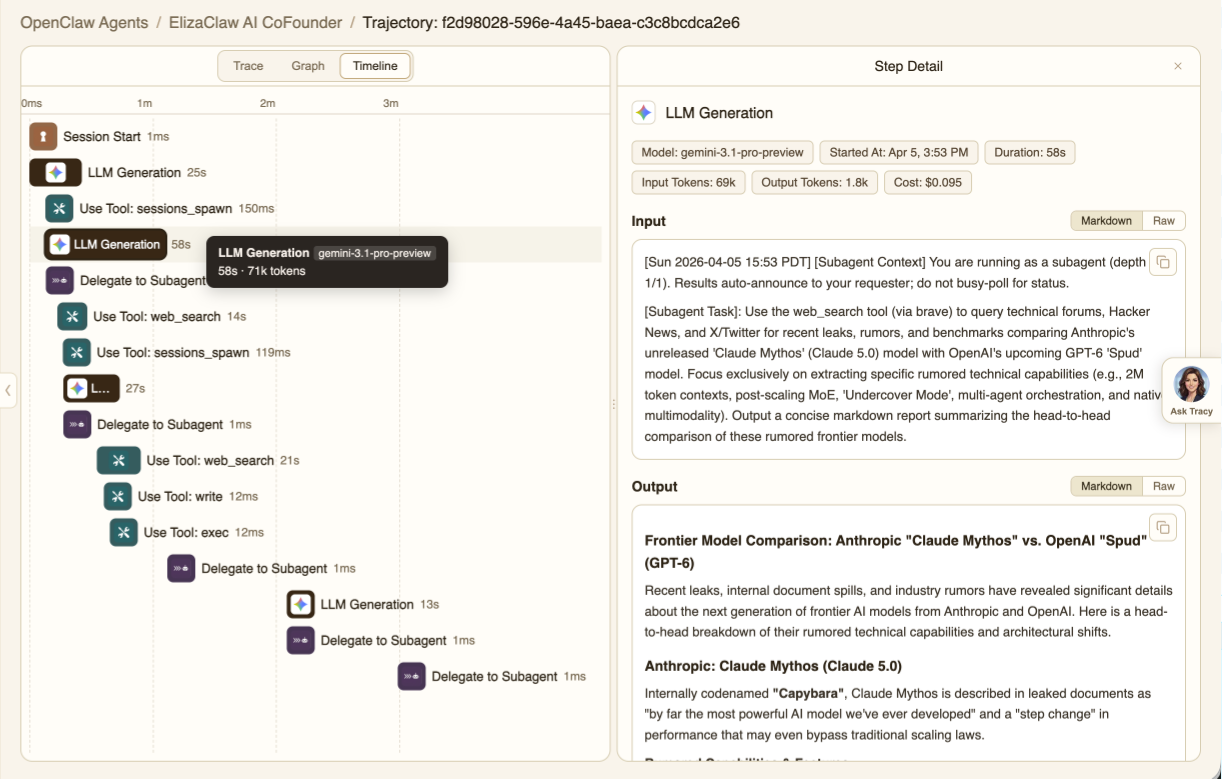}
  \caption{Step timeline (Gantt view) for a single trajectory.
  Each bar is one span; bar length is wall-clock duration.
  Redundant tool calls appear as visually identical adjacent bars,
  the same pattern the TraceCard \texttt{redundant\_tool\_calls}
  heuristic flags programmatically.}
  \Description{A ClawTrace timeline view for one agent trajectory. The left panel shows events placed along a minute-scale timeline, including session start, LLM generation spans, tool calls, web search calls, write and exec calls, and delegated sub-agent spans. A highlighted tooltip shows one LLM generation span with model name, duration, and token count. The right panel shows details for the selected LLM generation, including model, start time, duration, input tokens, output tokens, cost, input prompt text, and generated output text.}
  \label{fig:timeline}
\end{figure}

The trajectory dashboard (Figure~\ref{fig:dashboard}) lists all agent runs with columns for total cost, token count, step count, and outcome. Daily trend charts and per-agent filtering let operators spot cost spikes before drilling into individual traces. The step timeline (Figure~\ref{fig:timeline}) renders each span as a Gantt bar whose length is wall-clock duration, making parallelization opportunities and redundant call clusters visible at a glance.

Upcoming features include rubric-based evaluation, A/B testing across skill versions, and a closed loop that feeds trace-derived findings back into skill updates without manual intervention.

\subsection{Detailed Analysis of Findings}
\label{app:details}

\subsubsection{Why Cost Attribution Matters}

With the analyst, merge algorithm, and training set held fixed, stripping cost fields from TraceCards raises the median cost uplift on successful tasks from +22\% to +49\% and increases catastrophic no-deliverable failures from 1 to 5. The only variable that changed is what the analyst sees. Without per-step cost, the analyst removes structural features that were actually necessary, and the merge step does not catch the error until the agent fails to produce output. Five of the six No-cost-attribution regressions produce no deliverable at all ($Q{=}0$), a pattern Full CostCraft shows only once.

\subsubsection{Why Prune Is Protective Before Compressive}

Removing prune patches triples quality regressions (4 to 13) while median cost stays comparable (+15\% vs.\ +21\%). Eight of the 13 No-prune regressions produce no output: the agent runs tool calls but never writes the final answer. The Cost-control section of the skill, which contains only prune-derived rules, acts as a guardrail that keeps the other skill sections from breaking tasks that already worked. At this scale, the prune rules are protecting quality rather than compressing cost. The compression claim requires more data; the protection claim is what the current results support.
\begin{sloppypar}

\subsubsection{Sources of Quality Regression Under Full CostCraft}

Four quality regressions sit in the 30-task held-out set at seed 0. Task 42034 ($Q$: 1.0 $\to$ 0.667) regresses because an over-generalized placeholder-formatting repair patch writes \texttt{N/A} for missing values where the rubric expects empty cells. Task 48745 ($Q$: 1.0 $\to$ 0.667) regresses from an over-general preserve patch that misroutes a sheet-level write ordering. Task 55535 ($Q$: 1.0 $\to$ 0.471) follows a similar pattern; the No-prune skill preserves quality on this task because the regression-inducing rule lives in Workflow, not Cost-control. Task 76-42 ($Q$: 0.94 $\to$ 0.0) is the most severe of the four: the agent under Full CostCraft fails to produce a scorable deliverable. We did not isolate the responsible patch on this task.
\end{sloppypar}
\subsubsection{Why the Efficiency Regime Is Hard}

A prune rule can only help a held-out task whose baseline actually exhibits the waste pattern the rule targets. Agent behavior is stochastic: the same task run twice often wastes tokens on different things. The prune-rule library must grow with observed waste diversity, not just with training set size. At the 10-task training scale, the two learned prune rules match only 2 of 17 successful held-out tasks. On those tasks, Full CostCraft still increases cost (+38\% and +22\%), but the No-prune ablation causes one to fail entirely ($Q{=}0$). The prune rules are stabilizing quality on matched tasks rather than reducing their cost.

\subsubsection{Case Studies}
\label{app:cases}

\paragraph{Recovery: task 488-29 ($Q$: 0 $\to$ 1.0).} The baseline TraceCard logs a \texttt{failed\_or\_repaired\_step} at turn 4: the agent wrote a placeholder and ended the session without returning to fill it. The error analyst emitted a repair patch: ``When a cell is marked pending, compute it before ending the session.'' In the Full CostCraft run the agent computes and submits at $Q{=}1.0$. Cost rises from \$0.021 to \$0.118 (+461\%). The regime-partitioned evaluation makes this cost-for-correctness trade legible.

\paragraph{Prune protection: task 47484 ($Q{=}1.0$ preserved).} The baseline succeeds at $Q{=}1.0$ for \$0.068. Full CostCraft holds quality at 1.0 with cost \$0.083 (+22\%). Under No-prune, the agent produces no deliverable ($Q{=}0$, cost ${\approx}\$0$). The No-prune skill's Workflow section, stripped of Cost-control discipline, leads the agent into a terminal state without writing the answer sheet. Of the 8 No-prune catastrophic failures, 6 occur on tasks whose baselines succeed at $Q{=}1.0$.

\begin{sloppypar}

\paragraph{Global-mutation regression: task 42034 ($Q$: 1.0 $\to$ 0.667).} The Full CostCraft Workflow section adopts a preserved rule from a different trajectory (``Write \texttt{N/A} for missing placeholder values'') that conflicts with this task's rubric, which expects literal empty cells. This illustrates why we treat ``prune is protective'' as a statistical finding (13 vs.\ 4 regressions) rather than a per-task guarantee.
\end{sloppypar}

\subsubsection{SkillsBench Transfer Case Studies}
\label{app:sb-cases}

The five quality-changing events on the 84-task SkillsBench transfer benchmark fall into two clean groups by patch lane (Table~\ref{tab:skillsbench-events}). The two wins trace to the prune lane; the three regressions trace to the preserve lane. Per-task narratives below.

\paragraph{Win, prune lane: \protect\task{citation-check} ($Q$: 0.0 $\to$ 1.0).} The baseline run bills no LLM cost: the agent reads the input, immediately calls a tool that errors on a missing citation database connection, and exits without producing a scored deliverable. Under Full CostCraft, the ``read each file once and cache'' prune rule changes the access pattern: the agent reads the bibliography once, caches the result, and proceeds to the verification step that the baseline never reached. CostCraft cost is \$0.145; baseline cost is \$0. The win is not a cost-cheaper path but a path that exists at all.

\paragraph{Win, prune lane: \protect\task{glm-lake-mendota} ($Q$: 0.0 $\to$ 1.0).} The baseline costs \$0.233 and ends in failure: the agent re-reads the same Lake Mendota CSV file three times across the session, exhausts its turn budget on data reorganization, and never reaches the model-fitting step. Full CostCraft costs \$0.404 ($+74\%$) and reaches $Q{=}1.0$. Quality recovery comes at a cost premium because the prune rule frees enough headroom for the agent to attempt the substantive work, which is itself expensive on this task.

\paragraph{Regression, preserve lane: \protect\task{energy-market-pricing} ($Q$: 1.0 $\to$ 0.0, $-63\%$ cost).} The baseline succeeds at \$0.292; Full CostCraft fails at \$0.108. The merged skill's Workflow section contains a preserved rule from a SpreadsheetBench trajectory that instructs the agent to format outputs as a spreadsheet table. The SkillsBench verifier expects a JSON market-pricing report; the formatting rule routes the agent into a sheet-building path that bills less because it never produces the actual computation. Cheaper but unscored is the failure mode this paper warns about.

\paragraph{Regression, preserve lane: \protect\task{gravitational-wave-detection} ($Q$: 1.0 $\to$ 0.0, $-50\%$ cost).} Same mechanism as \protect\task{energy-market-pricing}. The baseline reads the strain time-series, runs a matched-filter step, and writes a detection table at \$0.531. Under Full CostCraft, the preserved formatting rules trigger an XLSX-style output path; the verifier expects a HDF5 metadata file plus a plain-text summary. Cost drops because the agent skips the heavy matched-filter step that the new output format does not require, and the verifier scores zero. The trace's \protect\task{top\_cost\_spans} confirms the matched-filter step is missing from the CostCraft trajectory.

\paragraph{Regression, preserve lane: \texttt{invoice-fraud-detection} ($Q$: 1.0 $\to$ 0.0, $+21\%$ cost).} The exception in the regression set: cost goes up rather than down. The agent under Full CostCraft applies the preserve-lane formatting rules and additionally runs the prune-lane cache discipline; the result is a longer session that produces a CSV-formatted answer instead of the expected JSON. Quality goes to zero because of the format mismatch; cost rises because the agent does the substantive analysis but then formats it incorrectly. This is the only regression where the agent reaches the verifier with a wrong-but-substantive answer rather than a no-deliverable trajectory.

The three preserve-lane regressions together motivate the rule-type-evaluation thesis. Each is a documented mechanism instance in which a SpreadsheetBench-specific convention misroutes the agent on a task type the convention does not apply to. None of them would have been caught by an aggregate cost or aggregate quality summary.

\subsubsection{Counterfactual Admission Ablation}
\label{app:cf}

On a 15-task success-heavy slice we compare Full CostCraft against No-counterfactual. The analyst emits counterfactuals voluntarily even with the gate disabled: all 3 prune patches under \protect\task{require\_counterfactual = False} contain a non-empty counterfactual field, because the LLM has internalized the prompt-level instruction independently of the admission check. Yet the merged skill regresses on 2 tasks that Full CostCraft preserves at $Q{=}1.0$. The gate appears to shape the quality of counterfactuals rather than their presence. The sample is small ($N{=}15$); we treat this as suggestive rather than conclusive.

\subsubsection{Tracing Overhead}
\label{app:overhead}

The ClawTrace plugin batches events in memory and flushes once per session in one HTTPS POST (typical payload 30 to 50\,kB). Over 20 trials the median HTTP round-trip is 445\,ms against a median agent trajectory wall time of 147\,s, yielding roughly 0.30\% overhead. A paired ON/OFF rerun on 10 SpreadsheetBench tasks showed zero quality divergence between instrumented and uninstrumented runs.

\subsubsection{Regime-Partitioned Breakdown}
\label{app:regime}

\begin{table}[h]
\centering
\small
\caption{Regime-partitioned breakdown (30 held-out tasks).}
\label{tab:regime}
\begin{tabular}{@{}llccc@{}}
\toprule
\textbf{Regime} & \textbf{Transition} & \textbf{N} & \textbf{Full CC} & \textbf{No-prune} \\
\midrule
Success (17) & s$\to$s preserved & 13 & $\Delta Q{=}0$ & 5 \\
 & s$\to$regress & 4 & --- & 12 \\
Partial (10) & tie / minor & 6 & $\Delta\${+}2\%$ & mix \\
 & recovery ($Q{\to}1.0$) & 3 & $\Delta Q{+}92$pp & 2 \\
 & partial$\downarrow$ & 1 & --- & 3 \\
Fail (3) & recovery ($Q{\geq}0.8$) & 2 & $\Delta Q{+}92$pp & 1 \\
 & fail$\to$fail & 1 & $\Delta\${+}193\%$ & 2 \\
\bottomrule
\end{tabular}
\end{table}

The three regimes show distinct cost-quality tradeoffs. In the \textbf{success regime} (17 tasks), Full CostCraft preserves quality on 13 tasks ($\Delta Q{=}0$) while increasing median cost by 22\%. The 4 regressions come from over-generalized preserve or repair patches that conflict with task-specific rubric requirements (Section~\ref{app:cases}). Under No-prune, 12 of 17 tasks regress, confirming that the Cost-control section is load-bearing for tasks that already pass.

In the \textbf{partial regime} (10 tasks, $0 < Q_{\mathrm{baseline}} < 1$), 3 tasks recover to $Q{=}1.0$ under Full CostCraft. These recoveries come from repair patches that address the failure modes identified in the operational taxonomy (Table~\ref{tab:taxonomy}): two are placeholder-mismatch fixes (T4), one is a no-deliverable fix (T1). The cost premium on recovered tasks averages +89\%, driven by additional LLM turns the agent uses to verify its output against the repair rule.

In the \textbf{fail regime} (3 tasks, $Q_{\mathrm{baseline}}{=}0$), 2 tasks recover to $Q{\geq}0.8$. The remaining task stays at $Q{=}0$ under all conditions; the error analyst could not diagnose its failure within the 3-lookup budget. Cost on this task rises by 193\% because the agent attempts all repair rules before giving up.

\subsubsection{Multi-Seed Robustness}
\label{app:multiseed}

% \begin{table}[h]
% \centering
% \small
% \caption{SpreadsheetBench held-out set (30 tasks) at \texttt{seed=0} and \texttt{seed=1}, plus pooled per-task means.}
% \label{tab:multiseed}
% \begin{tabular}{@{}lccc@{}}
% \toprule
% \textbf{Metric} & \textbf{Seed 0} & \textbf{Seed 1} & \textbf{Pooled} \\
% \midrule
% Full CC quality (mean / median)   & 0.85 / 1.00 & 0.89 / 1.00 & 0.87 / 1.00 \\
% No-prune quality (mean / median)  & 0.58 / 0.96 & 0.82 / 1.00 & 0.70 / 0.70 \\
% Full CC cost (mean / median)      & \$0.090 / \$0.089 & \$0.108 / \$0.100 & --- \\
% No-prune cost (mean / median)     & \$0.063 / \$0.073 & \$0.103 / \$0.092 & --- \\
% Full CC regressions               & 4 / 30 & 3 / 30 & 5 / 30 \\
% No-prune regressions              & 13 / 30 & 5 / 30 & 14 / 30 \\
% No-prune $Q{=}0$ (catastrophic)   & 11 / 30 & 1 / 30 & --- \\
% \bottomrule
% \end{tabular}
% \end{table}
\begin{table}[t]
\centering
\small
\caption{SpreadsheetBench held-out set (30 tasks) at \texttt{seed=0} and \texttt{seed=1}, plus pooled per-task means.}
\label{tab:multiseed}
\resizebox{\columnwidth}{!}{%
\begin{tabular}{@{}lccc@{}}
\toprule
\textbf{Metric} & \textbf{Seed 0} & \textbf{Seed 1} & \textbf{Pooled} \\
\midrule
Full CC quality (mean / median)   & 0.85 / 1.00 & 0.89 / 1.00 & 0.87 / 1.00 \\
No-prune quality (mean / median)  & 0.58 / 0.96 & 0.82 / 1.00 & 0.70 / 0.70 \\
Full CC cost (mean / median)      & \$0.090 / \$0.089 & \$0.108 / \$0.100 & --- \\
No-prune cost (mean / median)     & \$0.063 / \$0.073 & \$0.103 / \$0.092 & --- \\
Full CC regressions               & 4 / 30 & 3 / 30 & 5 / 30 \\
No-prune regressions              & 13 / 30 & 5 / 30 & 14 / 30 \\
No-prune $Q{=}0$ (catastrophic)   & 11 / 30 & 1 / 30 & --- \\
\bottomrule
\end{tabular}%
}
\end{table}

The regression-count gap between Full CC and No-prune is robust across seeds (factor $3.25\times$ at seed 0, $1.67\times$ at seed 1, $2.8\times$ pooled). The catastrophic ``no deliverable'' severity is more seed-sensitive: 11 of 13 No-prune regressions are $Q{=}0$ at seed 0, but only 1 of 5 at seed 1. Within-seed cost has a coefficient of variation near 43\% on this benchmark and backbone, so cost percentages reported in the main text are calibrated to two-seed pooled means; a third seed would not tighten them meaningfully.

\subsubsection{Cost Model}
\label{app:cost-model}

ClawTrace computes per-step cost as:
% \[
% \mathrm{cost} = r_{\mathrm{in}} \cdot t_{\mathrm{in}} + r_{\mathrm{out}} \cdot t_{\mathrm{out}} + r_{\mathrm{cacheRead}} \cdot t_{\mathrm{cacheRead}} + r_{\mathrm{cacheWrite}} \cdot t_{\mathrm{cacheWrite}}
% \]
\[
\begin{aligned}
\mathrm{cost} ={}&
r_{\mathrm{in}} \cdot t_{\mathrm{in}}
+ r_{\mathrm{out}} \cdot t_{\mathrm{out}} + r_{\mathrm{cacheRead}} \cdot t_{\mathrm{cacheRead}} \\
&+ r_{\mathrm{cacheWrite}} \cdot t_{\mathrm{cacheWrite}} .
\end{aligned}
\]
where $r_*$ are per-token USD rates and $t_*$ are token counts reported by the provider. For \texttt{openai-codex/gpt-5.4} (as of April 2026): $r_{\mathrm{in}}{=}\$2.00/\text{M}$, $r_{\mathrm{out}}{=}\$8.00/\text{M}$, $r_{\mathrm{cacheRead}}{=}\$0.50/\text{M}$, $r_{\mathrm{cacheWrite}}{=}\$2.00/\text{M}$. Cache-read tokens are billed at 25\% of the fresh input rate. On our 50-trajectory SpreadsheetBench sample, cache-read tokens constitute 30--50\% of total input volume. Counting them at the fresh input rate would overstate true cost by 1.6--2.0$\times$, distorting the step-level cost rankings that CostCraft relies on for prune-patch selection.

\subsection{CostCraft Prompt Templates}
\label{app:prompts}

This section reproduces the analyst and merge prompts used in CostCraft. All prompts are abbreviated; full versions are available in the code repository.
\newpage
\subsubsection{Success Analyst Prompt}

\begin{tcolorbox}[
  promptbox,
  colback=green!3!white,
  colframe=green!50!black,
  title={Success Analyst Prompt}
]
\textbf{Role:} You are an expert analyst for agent skill distillation.

\textbf{Mission:} Given a successful agent trajectory and its TraceCard, produce up to two skill patches: one \texttt{preserve} patch describing the behavior to keep, and optionally one \texttt{prune} patch targeting an expensive step that did not affect the outcome.

\textbf{Input:} (1) The TraceCard YAML for this session, including \texttt{top\_cost\_spans}, \texttt{redundant\_tool\_calls}, and \texttt{total\_cost\_usd}. (2) The current skill document (\texttt{SKILL.md}).

\textbf{Preserve patch requirements:}
\begin{itemize}[nosep,leftmargin=*]
\item Describe the behavior that contributed to the correct answer.
\item Phrase as a general principle, not a task-specific detail.
\item Reference the TraceCard step that demonstrates the behavior.
\end{itemize}

\textbf{Prune patch requirements (all three must hold):}
\begin{itemize}[nosep,leftmargin=*]
\item Name a specific entry from \texttt{top\_cost\_spans}.
\item Provide a natural-language counterfactual: why removing this step would not change the outcome.
\item Phrase the rule as a behavior to avoid, not a hard cost cap.
\end{itemize}

\textbf{Output format:} JSON with fields \texttt{action} (\texttt{preserve} or \texttt{prune}), \texttt{rule}, \texttt{target\_span} (prune only), \texttt{counterfactual} (prune only), \texttt{confidence} (high/medium/low).
\end{tcolorbox}

\subsubsection{Error Analyst Prompt}

\begin{tcolorbox}[
  promptbox,
  colback=red!3!white,
  colframe=red!50!black,
  title={Error Analyst Prompt}
]
\textbf{Role:} You are an expert failure-analysis agent for spreadsheet tasks.

\textbf{Mission:} Given a failed or partially failed agent trajectory, its TraceCard, and oracle access to the ground truth, diagnose the failure and emit a \texttt{repair} patch.

\textbf{Available tools} (budget: 3 calls total):
\begin{itemize}[nosep,leftmargin=*]
\item \texttt{inspect\_mismatches}: Read which rubric items the agent output failed to satisfy.
\item \texttt{read\_gold\_snippet}: Look up the expected cell values for a specific range.
\item \texttt{final\_patch}: Emit the repair patch and terminate.
\end{itemize}

\textbf{Required workflow:}
\begin{enumerate}[nosep,leftmargin=*]
\item Call \texttt{inspect\_mismatches} to identify the failure surface.
\item Trace the failure to a specific agent decision or missing step in the TraceCard.
\item If needed, call \texttt{read\_gold\_snippet} to confirm the expected output.
\item Call \texttt{final\_patch} with the repair rule grounded in the observed failure and oracle evidence.
\end{enumerate}

\textbf{Output format:} JSON with fields \texttt{action} (\texttt{repair}), \texttt{rule}, \texttt{failure\_type}, \texttt{evidence}, \texttt{confidence}.

\textbf{Constraint:} If you cannot diagnose the failure within 3 tool calls, emit a low-confidence patch. The merge step will deprioritize it.
\end{tcolorbox}
\subsubsection{Merge Operator Prompt}

\begin{tcolorbox}[
  promptbox,
  colback=blue!3!white,
  colframe=blue!50!black,
  title={Merge Operator Prompt}
]
\textbf{Role:} You are a skill-edit coordinator. You receive independently proposed patches from multiple analyst runs. Your job is to merge them into one coherent, non-redundant skill document.

\textbf{Priority order:}
\begin{enumerate}[nosep,leftmargin=*]
\item Repair patches with causal diagnosis (highest).
\item Prune patches with a named cost target and counterfactual.
\item Preserve patches that appear in two or more trajectories.
\item Singleton preserve patches (drop these).
\end{enumerate}

\textbf{Conflict resolution:} When two patches target the same behavior, repair supersedes prune, which supersedes preserve. When two patches of the same type conflict, keep the one with stronger evidence.

\textbf{Output structure:} The merged skill must have exactly five sections:
\begin{enumerate}[nosep,leftmargin=*]
\item \textbf{Trigger}: when the skill applies.
\item \textbf{Workflow}: step-by-step procedure (preserve-derived).
\item \textbf{Stop rules}: when to terminate (repair-derived).
\item \textbf{Artifact checklist}: output verification (repair-derived).
\item \textbf{Cost control}: behaviors to avoid (prune-derived).
\end{enumerate}

\textbf{Post-checks:} (1) All five section headings present. (2) Total length $\leq$ 1200 tokens. (3) No task-specific identifiers leak into the final skill.
\end{tcolorbox}
\subsubsection{Example TraceCard}
\begin{tcblisting}{
  promptbox,
  colback=gray!5!white,
  colframe=gray!60!black,
  title={Example TraceCard},
  listing only,
  listing options={
    basicstyle=\ttfamily\scriptsize,
    breaklines=true,
    breakatwhitespace=false,
    columns=fullflexible,
    keepspaces=true
  }
}
session_id: "sb-task-47484"
model: "openai-codex/gpt-5.4"
total_cost_usd: 0.068
total_tokens:
  input: 12840  output: 3210
  cacheRead: 8450  cacheWrite: 1200
outcome: "success"
top_cost_spans:
  - kind: llm  role_hint: tool_call
    cost_usd: 0.021
    tokens: {in: 4200, out: 890}
    args_sample: "read_file('input.xlsx')"
  - kind: llm  role_hint: tool_call
    cost_usd: 0.018
    tokens: {in: 3800, out: 720}
    args_sample: "read_file('input.xlsx')"
  - kind: llm  role_hint: final_reply
    cost_usd: 0.012
    tokens: {in: 2100, out: 980}
redundant_tool_calls:
  - cluster: ["span-3", "span-7"]
    tool: "read_file"  similarity: 0.94
sub_agents: []
failed_or_repaired: []
\end{tcblisting}

This TraceCard triggers the ``read each file once and cache'' prune rule: spans 1 and 2 both call \texttt{read\_file('input.xlsx')} with 94\% argument similarity, and the redundancy detector flags them as a cluster. The success analyst names span 2 as the prune target and argues that the second read returned byte-identical content, so skipping it would not change the outcome.

\end{document}